\documentclass[acmtog]{acmart}

\usepackage{multirow}

\AtBeginDocument{%
  }

\copyrightyear{2026}
\acmYear{2026}
\setcopyright{cc}
\setcctype{by}
\acmConference[SIGGRAPH Conference Papers '26]{Special Interest Group on Computer Graphics and Interactive Techniques Conference Conference Papers}{July 19--23, 2026}{Los Angeles, CA, USA}
\acmBooktitle{Special Interest Group on Computer Graphics and Interactive Techniques Conference Conference Papers (SIGGRAPH Conference Papers '26), July 19--23, 2026, Los Angeles, CA, USA}
\acmDOI{10.1145/3799902.3811215}
\acmISBN{979-8-4007-2554-8/2026/07}

\begin{document}

\title{CAdam: Context-Adaptive Moment Estimation for 3D Gaussian Densification in Generative Distillation}


\author{SeungJeh Chung}
\orcid{0009-0000-5306-6896}
\affiliation{%
 \institution{IIIXR Lab, Kyung Hee University}
 \city{Yongin}
 \country{Republic of Korea}}
\email{teclados078@khu.ac.kr}

\author{Geonho Park}
\orcid{0009-0005-7727-379X}
\affiliation{%
 \institution{IIIXR Lab, Korea University}
 \city{Seoul}
 \country{Republic of Korea}}
\email{rascal050902@korea.ac.kr}

\author{Misong Kim}
\orcid{0009-0009-3714-4685}
\affiliation{%
 \institution{IIIXR Lab, Kyung Hee University}
 \city{Yongin}
 \country{Republic of Korea}}
\email{misong023@khu.ac.kr}

\author{HyeongYeop Kang}
\orcid{0000-0001-5292-4342}
\affiliation{%
 \institution{IIIXR Lab, Korea University}
 \city{Seoul}
 \country{Republic of Korea}}
\email{siamiz_hkang@korea.ac.kr}

\renewcommand{\shortauthors}{Chung et al.}

\begin{abstract}
Adaptive densification is the engine of 3D Gaussian Splatting (3DGS). However, when transposed to the optimization-based Generative Distillation paradigm, this reconstruction-native mechanism reveals fundamental limitations, resulting in inefficient representations cluttered with redundant primitives. We diagnose this failure as a \textit{Densification Dilemma} stemming from the stochastic nature of generative guidance: the standard magnitude-based accumulation indiscriminately aggregates transient noise alongside geometric signals, making it difficult to strike a balance between over-densification and under-fitting. To resolve this, we introduce \textbf{Context-Adaptive Moment Estimation (CAdam)}, a novel framework that reinterprets densification as a statistically grounded signal verification problem. CAdam leverages the first moment of gradients to exploit the interference principle---where stochastic fluctuations cancel out via \textit{destructive interference} while consistent geometric drifts accumulate via \textit{constructive interference}---effectively disentangling the underlying signal from the generative noise floor. This is further augmented by a quantile-based context awareness and an intrinsic Signal-to-Noise Ratio (SNR) gating mechanism, which ensure robust adaptation across optimization stages and enable the soft termination of densification. Extensive experiments across diverse objectives (SDS, ISM, VFDS) and strong generative 3DGS backbones show that CAdam reduces Gaussian count by \textbf{85\%--97\%} relative to standard densification while preserving overall comparable perceptual quality. These results highlight signal-aware density control as a practical way to improve memory efficiency in optimization-based generative distillation.
\end{abstract}


\begin{teaserfigure}
  \includegraphics[width=\linewidth]{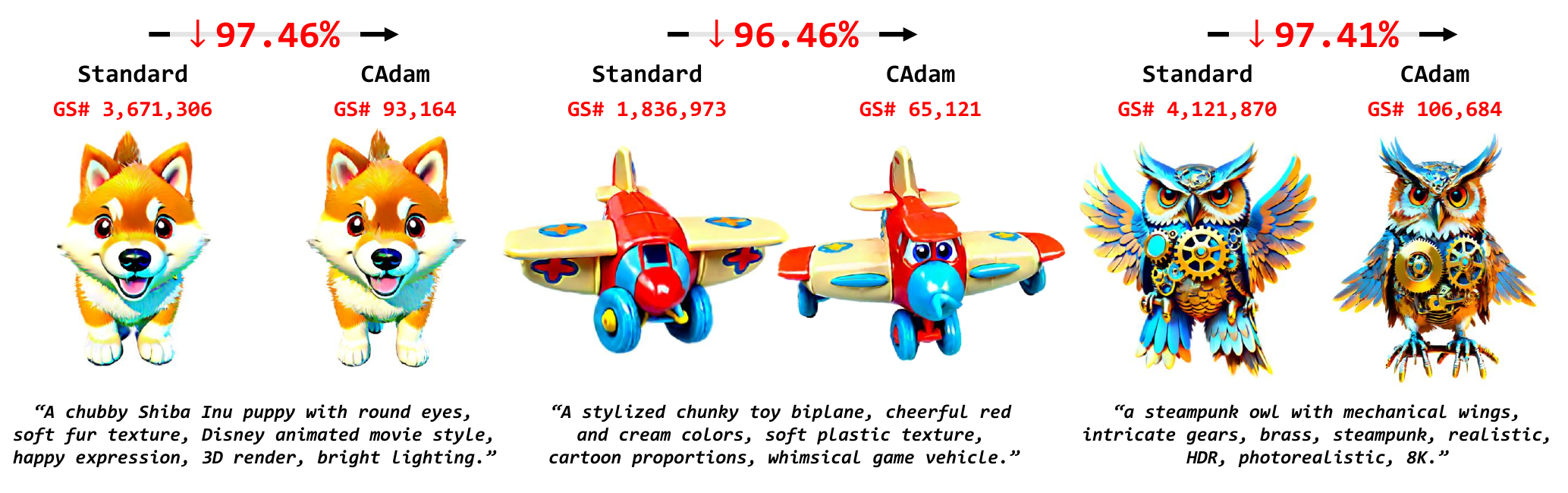}
  \caption{
  CAdam reduces Gaussian primitives by up to 97\% in optimization-based generative 3D Gaussian Splatting. Across diverse prompts and structures, our method achieves substantially more compact representations while maintaining overall comparable rendered appearance relative to standard densification.
  }
  \Description{Three side-by-side comparisons for a Shiba Inu puppy, a toy biplane, and a steampunk owl. Each pair compares standard densification and CAdam, showing that CAdam preserves a similar rendered appearance while reducing Gaussian counts by about 96 to 97 percent.}
  \label{fig:1_teaser}
\end{teaserfigure}

\maketitle

\section{Introduction}
\label{sec:intro}

Optimization-based generative 3D Gaussian Splatting (3DGS)~\cite{kerbl20233d} has rapidly emerged as a powerful paradigm for text-to-3D synthesis.
From early adaptations to recent high-performance systems, this optimization-based branch inherits the \textit{Generative Distillation} paradigm.
It ``dreams up'' 3D assets by distilling knowledge from pre-trained 2D diffusion~\cite{rombach2022high} or flow models~\cite{liu2022flow, lipman2022flow} into the 3D Gaussian space.
Despite substantial progress in initialization strategies and generative loss designs, the mechanism governing structural evolution---densification---remains largely inherited from reconstruction-oriented pipelines, without explicit adaptation for generative supervision.

Densification critically determines both representational efficiency and geometric fidelity.
However, existing generative 3DGS methods still rely on gradient-magnitude accumulation, a heuristic designed for reconstruction. 
This approach relies on a fundamental assumption: \textit{gradient magnitude is a reliable proxy for geometric error}. 
In reconstruction tasks, the optimization targets fixed ground-truth views. As the rendered images converge to this reference, the gradient magnitudes naturally diminish. Thus, persistent high magnitudes reliably indicate under-reconstruction.
In generative settings, however, supervision is provided by stochastic pseudo-targets induced by random timesteps, noise, and view sampling. As a result, gradients exhibit persistent non-convergence, even when the geometry is already plausible.
Under these dynamics, magnitude accumulation fails to distinguish coherent geometric drift from stochastic fluctuations, leading to a \textbf{Densification Dilemma}: low thresholds cause unbounded proliferation of redundant Gaussians, while high thresholds suppress valid refinement and induce geometric under-fitting.
This results in inefficient representations dominated by redundant primitives with little contribution to visual quality.

To resolve this dilemma, we introduce \textbf{Context-Adaptive Moment Estimation (CAdam)}, a densification framework that reinterprets densification not as a magnitude-driven heuristic, but as a statistical signal verification process.
Inspired by moment estimation in adaptive optimizers~\cite{kingma2015adam}, CAdam instills reliability into the densification process through three tightly coupled design principles.
First, \textbf{Momentum-based Signal Verification} accumulates gradient vectors instead of scalar magnitudes, allowing stochastic noise to cancel via destructive interference while preserving coherent geometric drift.
Second, \textbf{Context-Adaptive Selection} evaluates densification candidates using population-aware quantile ranking and an intrinsic Signal-to-Noise Ratio (SNR), ensuring that refinement is triggered only by statistically reliable signals.
Finally, \textbf{Selective Structural Refinement} restricts topological updates to verified candidates, preventing redundant primitive proliferation and enabling efficient capacity allocation.

CAdam generalizes across optimization-based generative 3DGS settings, including Score Distillation Sampling (SDS), Interval Score Matching (ISM), and Vector Field Distillation Sampling (VFDS).
Experiments demonstrate that CAdam reduces Gaussian counts by \textbf{85\%--97\%} compared to standard densification while maintaining overall comparable perceptual quality.
We provide a visual comparison of this structural efficiency in Fig.~\ref{fig:1_teaser}.

Our contributions are summarized as follows:
\begin{itemize}
    \item We diagnose a fundamental mismatch between reconstruction-oriented densification and the Generative Distillation paradigm, defining the \textbf{Densification Dilemma}: low thresholds trigger redundant primitive proliferation, whereas high thresholds suppress valid refinement.

    \item We propose \textbf{CAdam}, a statistically grounded density control framework. By verifying geometric signals via \textbf{Momentum} and employing a \textbf{Context-Adaptive Selection} strategy, we enable \textbf{Selective Structural Refinement} that strictly densifies only reliable regions.

    \item We demonstrate that CAdam serves as a unified solution across diverse generative 3DGS backbones. Extensive experiments confirm an 85\%--97\% reduction in Gaussian primitives while maintaining overall comparable perceptual quality, substantially reducing structural redundancy.
\end{itemize}

\section{Related Works}
\label{sec:related_works}

\subsection{3D Generation and Distillation}
\label{subsec:related_generation}

Text-to-3D distillation optimizes 3D representations using generative priors of 2D diffusion models. Early NeRF-based methods established SDS~\cite{poole2023dreamfusion, zhu2023hifa, wang2023score, yu2023text} and improved its diversity~\cite{wang2023prolificdreamer}, and stability~\cite{wu2024consistent3d, yan2025consistent}.

With the adoption of explicit primitives, 3DGS extended SDS to Gaussian representations~\cite{yi2024gaussiandreamer, tang2023dreamgaussian, chen2024text}. 
However, applying diffusion gradients independently across random noise levels exposes a limitation: lack of pathwise consistency, often leading to over-smoothed pseudo-targets and unstable multi-view optimization~\cite{alldieck2024score}.

To mitigate these issues, recent work has shifted toward trajectory-based distillation.
ODE-based methods leverage DDIM inversion \cite{liang2024luciddreamer} or latent consistency models~\cite{zhong2024dreamlcm}, while Flow Matching variants~\cite{li2024flowdreamer, miao2024dreamer, lukoianov2024score, li2025walking} construct deterministic trajectories that reduce the mean-seeking bias of standard SDS.
Complementary efforts address multi-view inconsistency and prior-induced bias through consistency objectives or 3D-aware regularization~\cite{li2024connecting, zhou2025consdreamer, jin2025debiasing, zhang2025improving, miao2025rethinking, zhu2025segmentdreamer, zhang2025structural}. In parallel, several extensions expand controllability and representation variants within generative 3DGS~\cite{cai2024dreammapping, he2024gvgen, di2025hyper, pham2024mvgaussian, huang2024placiddreamer, tran2024modedreamer, chen2024vividdreamer, zhuo2024vividdreamer}.

Orthogonal to recent feed-forward and native/structured-latent 3D methods targeting fast asset generation~\cite{li2025controllable, tang2025cycle3d, xu2024grm, zhang2024gs, tang2024lgm, liu2024novelgs, hu2025turbo3d, xiang2024structured, xiang2025trellis2, wu2025unilat3d, lai2025hunyuan3d}, we focus on optimization-based generative distillation. 
We observe that stochastic gradient fluctuations in such settings fundamentally hinder effective Gaussian densification.
Our work addresses this bottleneck with signal-aware densification that aligns primitive growth with the statistical properties of generative supervision.

\subsection{Densification and Structural Dynamics}
\label{subsec:related_densification}

Adaptive density control governs Gaussian cloning and splitting during 3DGS optimization. 
While effective in reconstruction, its heuristic nature often results in over- or under-densification, motivating refinements to densification criteria and operations.

Several works improve gradient-based densification by enforcing directional consistency or locally adaptive thresholds, leading to more stable density evolution~\cite{zhang2024pixel, ye2024absgs, fan2024trim}.
Others improve densification operations with more controllable and temporally stable update rules based on residual formulations or momentum-inspired mechanisms~\cite{kotovenko2025edgs, lyu2025resgs, yuan2025ema}. 
Other approaches incorporate additional cues, such as color and frequency information~\cite{kim2024color, li2025psrgs}, or explore stochastic and sparsity-oriented formulations, including probabilistic sampling and fragment-level pruning~\cite{kheradmand20243d, ye20243d}. Recent post-pruning methods further reduce redundant primitives after or alongside representation growth, using uncertainty scores or learnable pruning masks~\cite{zhang2024lp, hanson2025pup}.

All of these methods are developed under deterministic reconstruction assumptions. In contrast, we address densification in the stochastic setting of optimization-based generative distillation.

\section{Preliminaries}
\label{sec:preliminaries}

\subsection{Densification in 3D Gaussian Splatting}
\label{subsec:pre_densification}
3DGS framework~\cite{kerbl20233d} represents a scene as a set of anisotropic 3D Gaussians $G_i$, each parameterized by a mean position $\mu_i$, covariance $\Sigma_i$, opacity $\alpha_i$, and spherical harmonics (SH) coefficients. Since 3DGS begins with a deliberately coarse initialization for efficiency and stability, it relies on adaptive \textit{densification} to progressively
increase representational capacity. 

Densification candidates are identified through the accumulation of view-space positional gradients over a fixed window of $K$ optimization steps. Let $\mathbf{g}_{i,t} = \nabla_{\mu_i^{2D}} \mathcal{L}$ denote the view-space positional gradient of Gaussian $G_i$ at step $t$. The accumulated gradient statistic is defined as 
\begin{equation}
    \mathcal{A}_i = \frac{1}{K} \sum_{t=1}^{K} \| \mathbf{g}_{i,t} \|_2 \, ,
    \label{eq:accum_metric}
\end{equation}
which serves as a proxy for persistent positional instability.
If $\mathcal{A}_i$ exceeds a predefined threshold $\tau_{pos}$, the Gaussian is selected for densification. 
The specific densification operation is determined by its scale $S_i$:
\begin{equation}
    \text{Action}(G_i) = 
    \begin{cases} 
    \text{Split}, & \text{if } \mathcal{A}_i > \tau_{pos} \text{ and } \|S_i\|_\infty > \tau_{scale} \\
    \text{Clone}, & \text{if } \mathcal{A}_i > \tau_{pos} \text{ and } \|S_i\|_\infty \leq \tau_{scale}
    \end{cases} \, .
\end{equation}

To maintain numerical stability and prevent degenerate primitives, Gaussians with negligible opacity or excessively large scales are subsequently pruned.

This strategy relies on a fundamental assumption of reconstruction: the positional gradient norm is a reliable indicator of the \textit{photometric error}:
\begin{equation}
    \mathcal{L}_{photo} = (1 - \lambda) \| \hat{I} - I_{GT} \|_1 + \lambda \text{D-SSIM}(\hat{I}, I_{GT}) \, .
    \label{eq:photo_error}
\end{equation}
where $\hat{I}$ is a rendered image and $I_{GT}$ is a ground-truth image. As reconstruction progresses, gradients for well-represented regions naturally decay, while persistently large gradients reliably indicate under-reconstruction.

\subsection{Gradient Formulation in Generative Distillation}
\label{subsec:pre_generation}
In contrast to reconstruction, text-to-3D generation optimizes a 3D representation $\theta$ by distilling knowledge from a pretrained 2D generative prior.
Most distillation-based objectives, ranging from Score Distillation Sampling (SDS) to recent advances like Interval Score Matching (ISM) and Vector Field Distillation Sampling (VFDS), can be unified into a general \textit{pseudo-target alignment} framework. 

Specifically, the update direction aligns the latent representation $x_0 := g(\theta, c)$ of $\hat{I}$ with the denoised target $\hat{x}_0^t$ estimated from the noisy latent $x_t$. 
Here, $c$ denotes the rendering conditions, including the camera pose and intrinsic parameters under which the image is rendered.
Incorporating a timestep-dependent weighting function $w(t)$ and the domain scaling factor $\gamma(t) = \frac{\sqrt{1 - \bar{\alpha}_t}}{\sqrt{\bar{\alpha}_t}}$, the gradient is formulated as:
\begin{equation}
    \nabla_\theta \mathcal{L}_{\text{SDS}} \approx \mathbb{E}_{t, \epsilon, c} \left[ \frac{w(t)}{\gamma(t)} ( x_0 - \hat{x}_0^t ) \frac{\partial g(\theta, c)}{\partial \theta} \right] \, .
    \label{eq:sds_gradient}
\end{equation}

Crucially, the discrepancy term $(x_0 - \hat{x}_0^t)$ admits equivalent parameterizations under linear diffusion dynamics~\cite{li2025back}:
\begin{equation}
    (x_0 - \hat{x}_0^t) \propto (\hat{\epsilon}_\phi - \epsilon) \propto (\hat{v}_\phi - v_t) \, .
\end{equation}
where $\hat{\epsilon}_\phi$ and $\hat{v}_\phi$ represent the noise and velocity predicted by the pre-trained model, respectively. This equivalence demonstrates that various guidance strategies share a common objective: minimizing the error between the current state and a predicted target on the learned data manifold. 

However, unlike reconstruction, this supervision is inherently \textit{stochastic}. Each update relies on single-step denoising at randomly sampled timesteps and noise realizations, producing gradients with high variance and no notion of convergence to a fixed target. This stochasticity persists even in interval-based variants and is further amplified by view sampling and manifold misalignment during early optimization. As a result, gradient magnitudes fluctuate persistently over time, violating the assumptions under which reconstruction-driven densification remains reliable.

\begin{figure}[t]
    \centering
    \includegraphics[width=\linewidth]{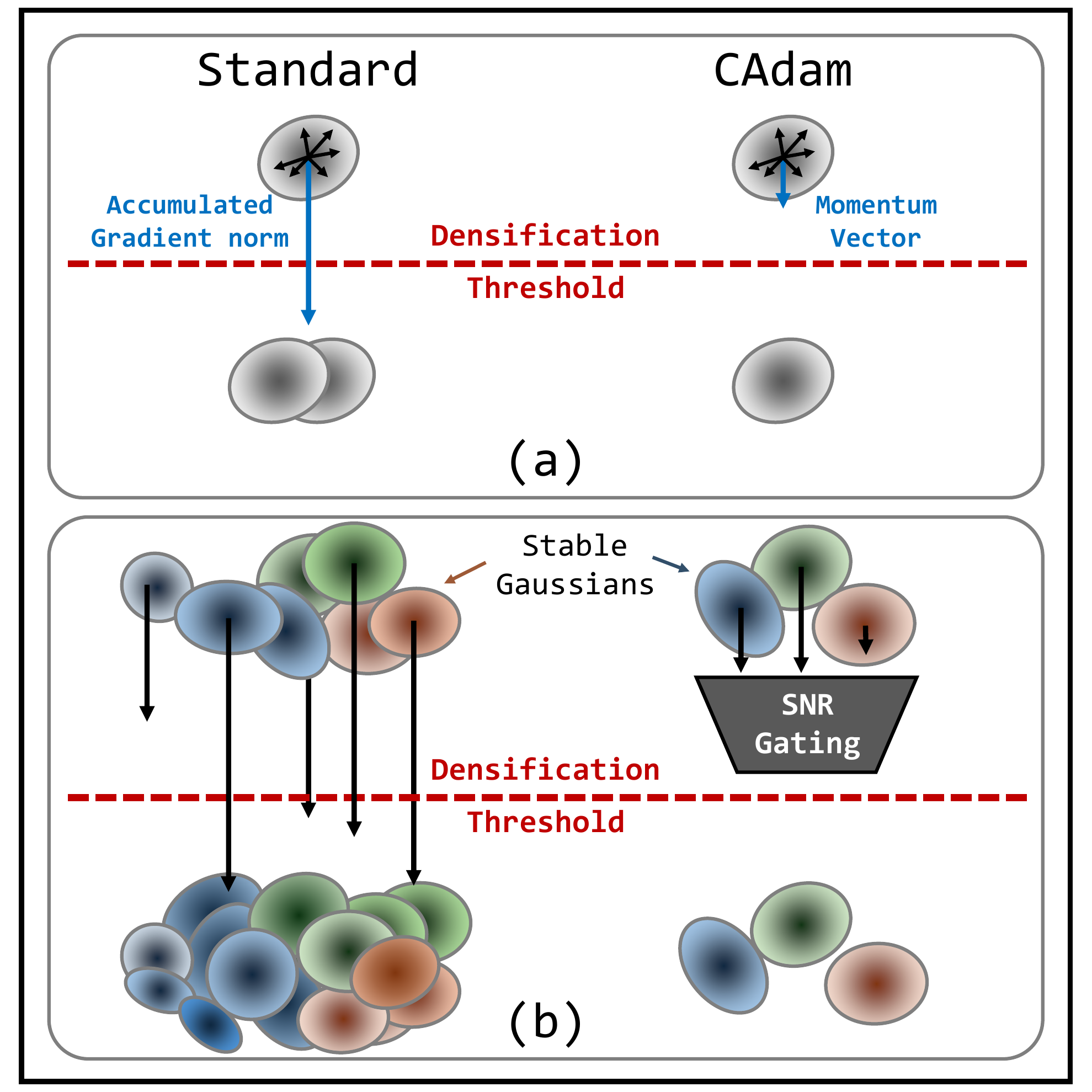}
    \caption{Conceptual comparison between magnitude-based densification and CAdam. (a) Standard densification uses the accumulated gradient norm as a scalar trigger, whereas CAdam relies on the momentum vector. (b) For geometrically stable Gaussians, standard accumulation repeatedly exceeds the densification threshold, leading to unnecessary splitting. CAdam suppresses this behavior through SNR gating.}
    \Description{Two schematic panels comparing standard densification and CAdam. Standard densification uses accumulated gradient norms and can repeatedly exceed the densification threshold for stable Gaussians, whereas CAdam uses a momentum vector and SNR gating to suppress unnecessary splitting.}
    \label{fig:2_pipeline}
\end{figure}

\section{Method}
\label{sec:method}
We introduce \textbf{CAdam}, a densification framework that reinterprets 3D Gaussian densification as a problem of statistical signal verification, rather than heuristic accumulation of gradient magnitudes.

Before detailing CAdam, we formalize why the standard densification mechanism---originally designed for reconstruction---
leads to a fundamental failure mode in generative densification.

\subsection{Problem Definition}
\label{subsec:method_problem}
Standard densification strategy relies on the accumulation of positional gradient magnitudes, implicitly assuming that gradients decay as optimization progresses. While valid for reconstruction, we demonstrate that it is fundamentally violated in generative distillation due to its non-convergent and stochastic supervision. 

\paragraph{Convergence in Reconstruction}
In reconstruction, optimization is driven by a static objective defined over fixed ground-truth views. 
The rendered image $\hat{I}$ converges to the target image $I_{GT}$, and the photometric loss $\mathcal{L}_{photo}$ naturally diminishes. Consequently, the gradients decay over time ($\lim_{t \to \infty} \| \nabla_{\mu} \mathcal{L} \| = 0$). Although random view sampling introduces mild stochasticity, the gradients still persistently point toward a single optimum. 
In this regime, accumulating large gradient norms reliably identifies regions of geometric under-fitting. 

\paragraph{Instability in Generative Distillation}
Generative distillation replaces the static ground truth with a stochastic pseudo-target $\hat{x}_0^t$, obtained via single-step denoising at randomly sampled timesteps $t$ and noise realizations $\epsilon$.
The optimization target therefore changes at every iteration, including two distinct dynamics:
(1) \textbf{Non-convergence:} The alignment error $(x_0 - \hat{x}_0^t)$ remains inherently non-zero, resulting in persistent gradient magnitudes even when the geometry is visually plausible.
(2) \textbf{Directional Variance:} Variance in the noise prediction $\hat{\epsilon}_\phi$ causes rapid fluctuations in the update direction. We refer to this high-frequency oscillatory behavior---where stochastic guidance yields conflicting gradient signals across steps---as the \textit{zigzag phenomenon}.

\paragraph{Densification Dilemma}
The failure arises when reconstruction-oriented densification is applied under these stochastic dynamics.
Let $\mathbf{g}_t$ denote the gradient of each Gaussian $G_i$ at step $t$, modeled as the superposition of a deterministic geometric drift $\bar{\mathbf{g}}_t$ and a stochastic noise component $\boldsymbol{\xi}_t$:
\begin{equation}
    \mathbf{g}_t = \bar{\mathbf{g}}_t + \boldsymbol{\xi}_t, \quad \text{where } \mathbb{E}[\boldsymbol{\xi}_t] = \mathbf{0} \, ,
\end{equation}
where $\bar{\mathbf{g}}_t$ represents the coherent update direction for structural refinement and $\boldsymbol{\xi}_t$ captures the stochastic fluctuations driving the \textit{zigzag phenomenon}. 

Ideally, densification should respond only to $\|\bar{\mathbf{g}}_t\|$.  However, due to the convexity of the $L_2$ norm, the standard metric inherently entangles the signal with the noise magnitude:
\begin{equation}
    \mathcal{A}_{naive} = \frac{1}{K} \sum_{t=1}^{K} \| \bar{\mathbf{g}}_t + \boldsymbol{\xi}_t \| \approx \| \bar{\mathbf{g}}_t \| + \frac{1}{K} \sum_{t=1}^{K} \| \boldsymbol{\xi}_t \| \, .
\end{equation}
Even when the geometry is plausible ($\bar{\mathbf{g}}_t \approx \mathbf{0}$), the accumulated noise floor $\frac{1}{K} \sum \| \boldsymbol{\xi}_t \|$ remains significantly non-zero. 

This leads to a \textbf{densification dilemma}: a low threshold captures fine details but is dominated by noise, triggering unbounded proliferation of redundant Gaussians; a high threshold suppresses noise but prevents valid refinement, yielding coarse or incomplete geometry. Crucially, no fixed threshold can resolve this ambiguity.

To resolve this limitation, we must shift densification paradigm from ``magnitude-based detection'' to ``statistical signal verification'', via moment estimation to cancel out $\boldsymbol{\xi}_t$ and recover $\bar{\mathbf{g}}_t$.

\subsection{Momentum-based Signal Verification}
\label{subsec:method_momentum}
To resolve the densification dilemma, we interpret momentum not as an optimization aid, but as a mechanism for verifying coherent geometric signals (see Fig.~\ref{fig:2_pipeline}). We leverage the first-moment estimate from Adam~\cite{kingma2015adam} to disentangle the coherent geometric drift $\bar{\mathbf{g}}_t$ from the stochastic noise $\boldsymbol{\xi}_t$. 

Unlike magnitude-based accumulation, we track positional gradients directly in world space ($\mathbf{g}_t = \nabla_{\mu_{xyz}} \mathcal{L}$), removing projection-dependent fluctuations from random camera sampling. The momentum vector $\mathbf{m}_t$ is updated via an exponential moving average (EMA) and bias-corrected to obtain an unbiased estimate $\hat{\mathbf{m}}_t$:
\begin{equation}
    \mathbf{m}_t = \beta_1 \mathbf{m}_{t-1} + (1 - \beta_1) \mathbf{g}_t, \quad \hat{\mathbf{m}}_t = \frac{\mathbf{m}_t}{1 - \beta_1^t} \, ,
    \label{eq:momentum_update}
\end{equation}
where $\beta_1$ is the decay rate. For newly instantiated Gaussians, we use each Gaussian's age as the step count $t$ for bias correction, rather than the global iteration count.

This vector-based integration exploits a key property of generative guidance: stochastic noise is zero-mean ($\mathbb{E}[\boldsymbol{\xi}_t] = \mathbf{0}$). Then, high-frequency fluctuations ($\boldsymbol{\xi}_t$) undergo \textit{destructive interference} and cancel out towards zero. Conversely, the consistent geometric drift $\bar{\mathbf{g}}_t$ accumulates via \textit{constructive interference}, resulting in a stable direction vector.

Consequently, the unbiased momentum effectively extracts the underlying geometric signal ($\hat{\mathbf{m}}_t \approx \bar{\mathbf{g}}_t$), yielding a robust densification criterion. Even in fine-detail regions where gradient magnitudes are small, directional consistency is preserved, allowing valid refinement to be detected where magnitude-based heuristics fail.

\subsection{Context-Adaptive Selection}
\label{subsec:method_SNR}
Identifying valid densification candidates requires more than the absolute magnitude of momentum $\hat{\mathbf{m}}_i$. As gradient scales decay over optimization and generative guidance remains stochastic, we propose a \textbf{Context-Adaptive Selection} strategy that jointly considers \textit{relative importance} within the scene and \textit{absolute reliability} of each signal.

\paragraph{Quantile-based Selection}
In response to shifting gradient scales, we adaptively select Gaussians for densification based on their relative importance within the scene. A dynamic threshold derived from population statistics is employed to identify candidate $G_i$. Specifically, $G_i$ is considered for densification if it satisfies:
\begin{equation}
    \|\hat{\mathbf{m}}_i\| > \text{Quantile}\left( \{ \| \hat{\mathbf{m}}_j \| \}_{j=1}^N, \tau_Q \right) \, ,
\end{equation}
where $\tau_Q$ is the target percentile (e.g., 0.9). 

This strategy remains effective throughout optimization. 
In early stages, it prioritizes coarse structural corrections with large momentum, while in later stages--when global gradients diminish--it continues to surface fine-detail regions whose momentum, though small in absolute value, remains significant compared to stabilized areas.
This quantile-based approach naturally induces an automated \textit{coarse-to-fine} refinement schedule without requiring a hand-tuned threshold. 

\paragraph{SNR-based Selection}
Momentum identifies directional trends but does not quantify signal confidence. To verify whether a candidate represents a reliable geometric signal rather than accumulated noise, we introduce a \textbf{reliability condition} based on the Signal-to-Noise Ratio (SNR). We leverage the second raw moment $\mathbf{v}_t$ from Adam~\cite{kingma2015adam}:
\begin{equation}
    \mathbf{v}_t = \beta_2 \mathbf{v}_{t-1} + (1-\beta_2)(\mathbf{g}_t \odot \mathbf{g}_t), \quad \hat{\mathbf{v}}_t = \frac{\mathbf{v}_t}{1 - \beta_2^t} \, .
\end{equation}

We define the Intrinsic SNR as
\begin{equation}
    \text{SNR}_t = \frac{\| \hat{\mathbf{m}}_t \|_2}{\sqrt{\| \hat{\mathbf{v}}_t \|_1} + \epsilon} \, .
\end{equation}
measuring the strength of coherent drift relative to stochastic fluctuations. A Gaussian is selected for densification only if $\text{SNR}_t > \tau_{\text{SNR}}$.

This metric effectively distinguishes signal from noise; Gaussians $G_i$ dominated by stochastic fluctuations exhibit small $\|\hat{\mathbf{m}}_i\|$ due to \textit{destructive interference} but large $\|\hat{\mathbf{v}}_i\|$ due to magnitude accumulation, resulting in a low SNR.

Densification is finalized by the intersection of quantile-based ranking and SNR criteria. As geometry converges ($\bar{\mathbf{g}}_t \to \mathbf{0}$), the coherent signal in the momentum vanishes, whereas stochastic fluctuations continue to accumulate in the variance term $\hat{\mathbf{v}}_t$. This naturally drives the SNR below $\tau_{\text{SNR}}$ even for top-quantile Gaussians. This intersection acts as a \textbf{soft termination mechanism}, ceasing densification once no statistically significant geometric drift remains and preventing redundant primitives.

\subsection{Selective Structural Refinement}
\label{subsec:method_refinement}

We decouple \textit{when} to densify from \textit{how} to modify geometry. The densification trigger is governed solely by statistical consistency (Momentum and SNR), while the structural operation (Clone vs. Split) follows the standard spatial scale criterion.

\paragraph{Action Criteria}
Once a Gaussian is selected for densification ($M_{\text{densify}} = \text{True}$), the structural adjustment is determined as follows:
\begin{equation}
    \text{Action} = 
    \begin{cases} 
    \text{Split}, & \text{if } \|S_i\|_\infty > \tau_{\text{scale}} \\
    \text{Clone}, & \text{otherwise}
    \end{cases}
\end{equation}
This separation ensures that temporal signal verification is orthogonal to spatial resolution control, allowing CAdam to be seamlessly integrated with existing 3DGS structural heuristics without introducing new geometric hyperparameters.

\paragraph{Selective Opacity Reset}
The standard practice of periodically resetting global opacity imposes a dilemma: it indiscriminately degrades plausible geometries and introduces artificial gradient spikes that corrupt momentum history (EMA) with non-geometric transients.
To mitigate this, we introduce a \textbf{Selective Opacity Reset}, targeting only unreliable primitives identified by low Intrinsic SNR after an initial warm-up:
\begin{equation}
    \alpha_{new} = 
    \begin{cases} 
    0.01, & \text{if } \text{SNR} < \tau_{\text{SNR}} \\
    \alpha_{old}, & \text{otherwise}
    \end{cases}
\end{equation}

Noise-dominated primitives naturally exhibit low SNR and are selectively suppressed, facilitating their removal during pruning.
This targeted reset ensures a balance between stability (preserving established structures) and plasticity (removing noisy artifacts to free up capacity for valid geometry), maintaining the statistical integrity of momentum statistics throughout training.

\begin{figure*}[t]
    \centering
    \includegraphics[width=\linewidth]{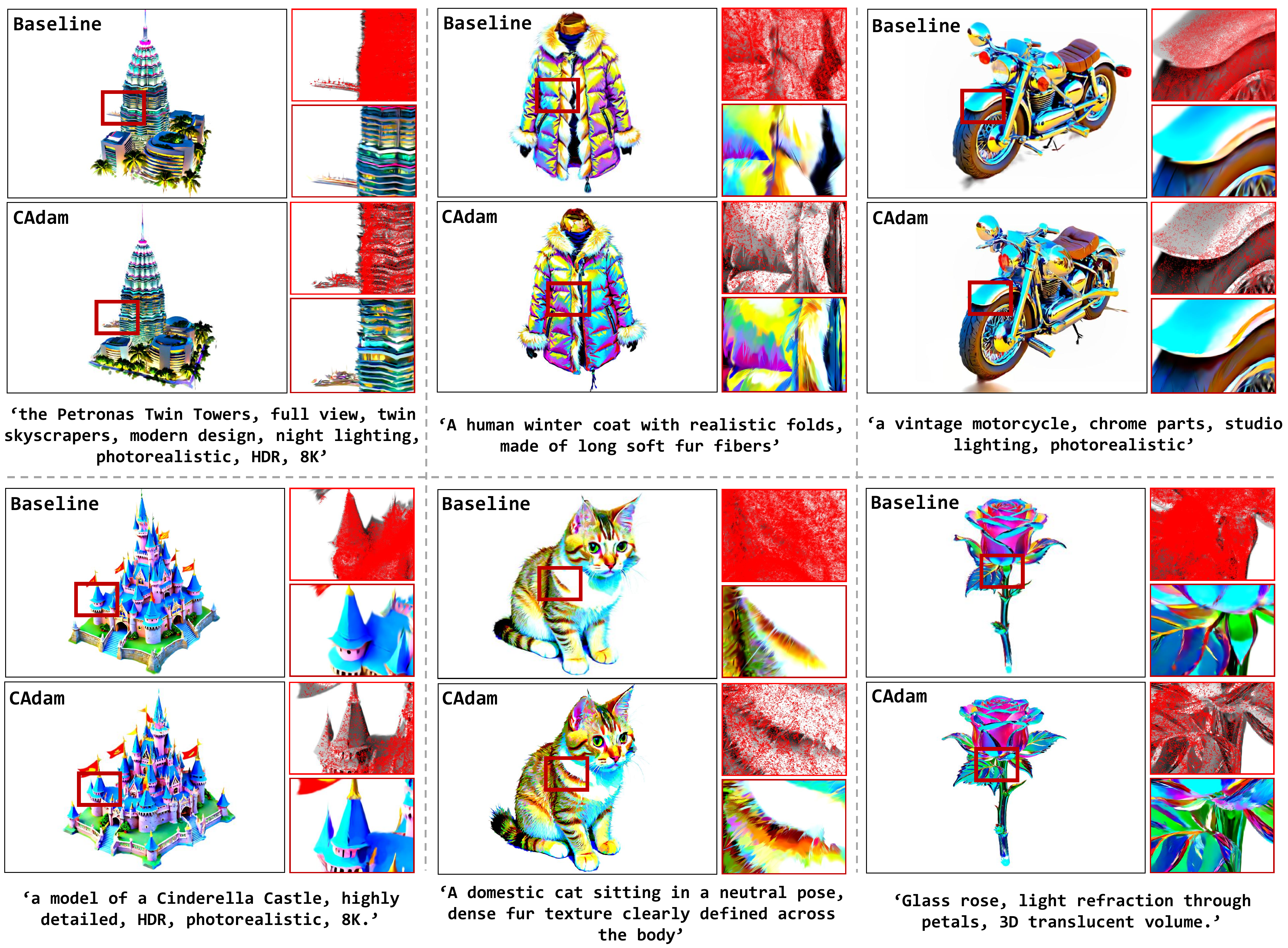}
    \caption{Qualitative comparison between standard densification (Baseline) and CAdam across diverse prompts and structures. Zoomed-in regions highlight that baseline densification produces smoothed, blurry overlap patterns, whereas CAdam yields cleaner, more compact representations.}
    \Description{Six qualitative comparisons for the Petronas Twin Towers, a winter coat, a motorcycle, a Cinderella Castle, a domestic cat, and a glass rose. Each example compares baseline densification and CAdam with zoomed-in regions, showing cleaner local structure and fewer redundant patterns with CAdam.}
    \label{fig:3_ex_our_results}
\end{figure*}

\section{Experiments}
\label{sec:experiments}

\begin{figure*}[t]
    \centering
    \includegraphics[width=\linewidth]{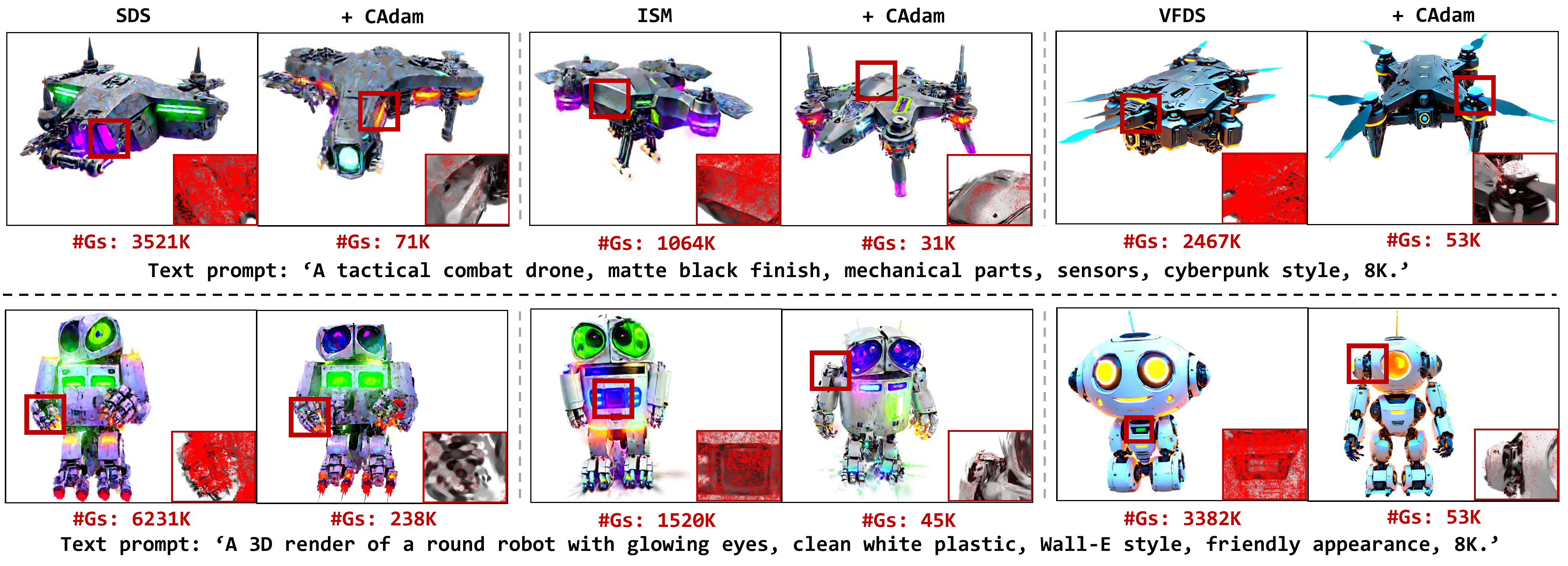}
    \caption{Generalization across distillation objectives. CAdam consistently reduces Gaussian primitives across SDS, ISM, and VFDS. Insets highlight that efficiency gains stem from removing representational redundancy rather than degrading geometric detail.}
    \Description{Two prompt examples comparing SDS, ISM, and VFDS objectives with and without CAdam. The drone and robot examples show that CAdam greatly reduces Gaussian counts across objectives while preserving the main object shape and local details.}
    \label{fig:4_ex_generalization1}
\end{figure*}

\subsection{Experimental Setup}
\label{subsec:exp_setup}

\paragraph{Baseline and Implementation}
We compare CAdam against the standard 3DGS densification~\cite{kerbl20233d}, which is the default density control used by most generative pipelines.
To assess generality, CAdam is integrated into multiple representative backbones and distillation objectives.
For controlled analyses of optimization behavior and ablations, we use a representative optimization-based 3DGS pipeline with VFDS loss and Stable Diffusion 3.5.
All experiments are run on a single NVIDIA RTX A5000 (24GB). Unless specified, we use a fixed set of hyperparameters for CAdam to demonstrate robustness: quantile threshold $\tau_Q = 0.9$, SNR threshold $\tau_{\text{SNR}} = 0.1$, and momentum decay $\beta_1 = 0.9$.
CAdam keeps each backbone's original $K$-step densification schedule: moment/SNR states are updated online at every optimization step, while split/clone candidates are selected only at scheduled densification calls using the quantile--SNR gate. The selective opacity reset schedule is also retained. See the supplementary material for implementation details and pseudocode.

\paragraph{Evaluation Protocol}
We evaluate CAdam on diverse text prompts from established benchmarks such as T3Bench~\cite{he2023t3bench} and the DreamFusion Gallery~\cite{poole2023dreamfusion}, supplemented by custom prompts generated via Gemini-Pro and GPT-4.
From an initial pool of 400 prompts spanning three categories of \textit{Material} (e.g., fur, glass), \textit{Structure} (e.g., complex architecture), and \textit{Concept} (e.g., fantasy scenes),  we sample 50 prompts with a balanced distribution.

We organize evaluation along two complementary axes.
First, we assess \textit{visual fidelity} using three criteria: 
(1) \textbf{Semantic Alignment}, measured by CLIP Score~\cite{hessel2021clipscore} and ImageReward~\cite{xu2023imagereward}, for prompt consistency;
(2) \textbf{Perceptual Aesthetics}, evaluated via HPS v2~\cite{wu2023human}, reflecting human visual preference; and
(3) \textbf{Cleanliness and Naturalness}, assessed by NIQE~\cite{mittal2012making}, where lower values indicate fewer visual artifacts and noise. 
Second, we evaluate \textit{efficiency}, measured by the \textit{Number of Gaussians ($N_g$)} and \textit{Storage Size} (MB). To complement the evaluation with geometric evidence, depth and multi-view visualizations are provided in the supplementary material.

\subsection{Comparative Evaluation}
\label{subsec:exp_compare}
We compare CAdam against standard densification strategy used in optimization-based generative 3DGS pipelines.
As shown in Fig.~\ref{fig:3_ex_our_results}, CAdam produces visually detailed 3D assets while using substantially fewer primitives than the baseline.

\begin{figure*}[p]
    \centering
    \includegraphics[width=.96\linewidth]{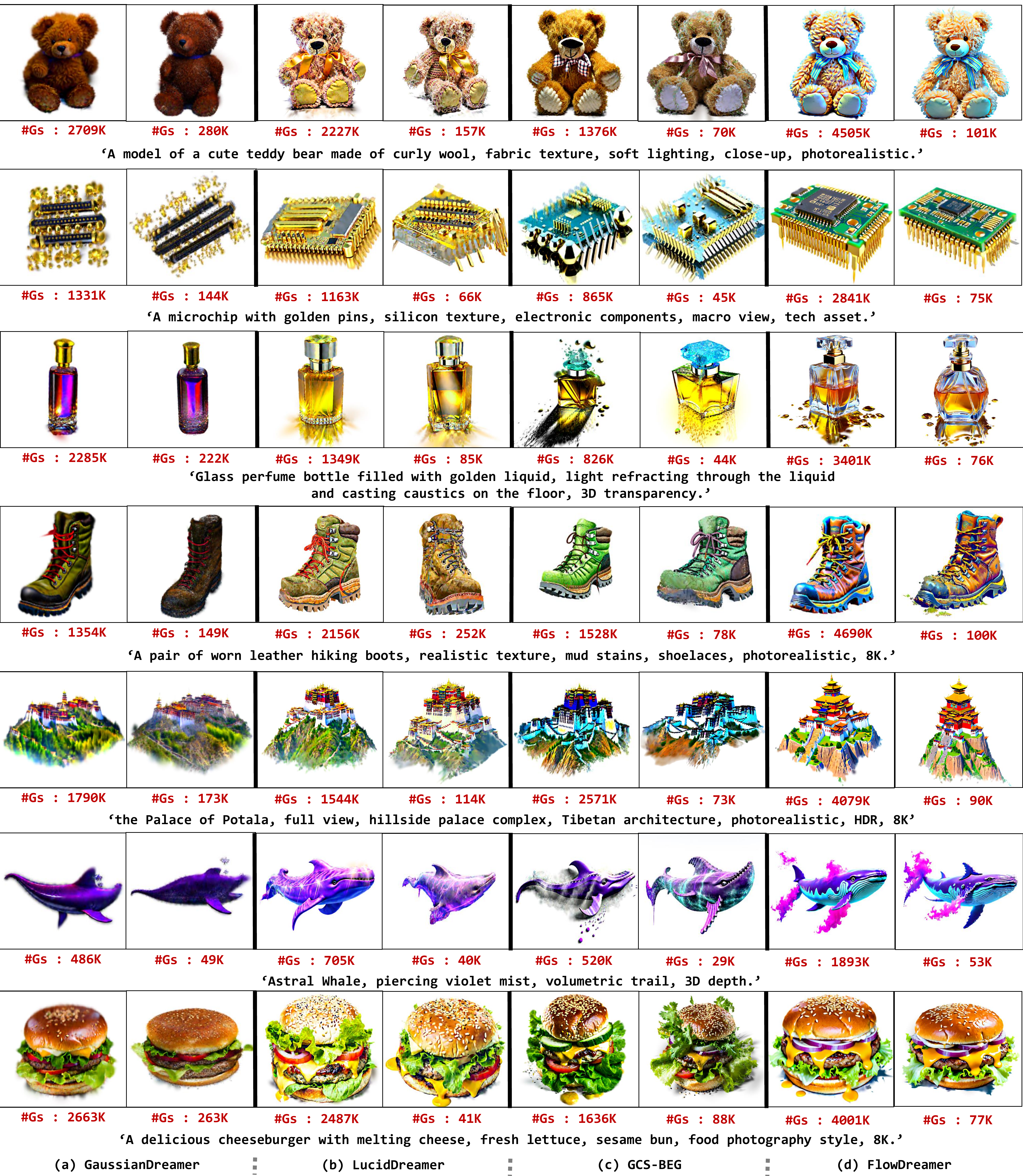}
    \caption{Model-agnostic generalization of CAdam across optimization-based generative 3DGS frameworks. Qualitative comparisons on four generative backbones show that CAdam maintains comparable visual fidelity while dramatically reducing Gaussian counts. CAdam's efficiency gains are robust and independent of the underlying backbone.}
    \Description{A grid of results across four generative 3D Gaussian Splatting backbones: GaussianDreamer, LucidDreamer, GCS-BEG, and FlowDreamer. Rows show prompts such as a teddy bear, a microchip, a perfume bottle, hiking boots, Potala Palace, an astral whale, and a cheeseburger, with Gaussian counts reported for each comparison.}
    \label{fig:5_ex_generalization2}
\end{figure*}

\paragraph{Generalization and Visual Fidelity}
We validate the generalizability of CAdam across different distillation objectives and generative backbones.
Fig.~\ref{fig:4_ex_generalization1} shows that CAdam consistently improves efficiency under SDS, ISM, and VFDS.
Fig.~\ref{fig:5_ex_generalization2} further demonstrates consistent behavior across  GaussianDreamer~\cite{yi2024gaussiandreamer}, LucidDreamer~\cite{liang2024luciddreamer}, GCS-BEG~\cite{li2024connecting}, and FlowDreamer~\cite{li2024flowdreamer}. 
Across all methods, CAdam significantly reduces primitive counts while maintaining comparable visual fidelity, demonstrating that its gains are model-agnostic.

\paragraph{Quantitative Compactness and Fidelity}
Tab.~\ref{tab:quantitative} summarizes performance across GaussianDreamer (SDS), LucidDreamer (ISM), and FlowDreamer (VFDS). 
Across all backbones, CAdam significantly reduces the number of Gaussians by \textbf{85\%--97\%}, yielding proportional storage savings compared to standard densification. 
Despite this substantial compression, quality metrics remain broadly comparable: CLIP and NIQE fluctuate across settings, while ImageReward improves under all objectives and HPS v2 remains comparable. 
These results support CAdam as a compactness-oriented plug-and-play densification module that substantially reduces redundancy while maintaining broadly comparable quality.

\begin{table}[t]
    \centering
    \renewcommand{\arraystretch}{1.2}
    \setlength{\tabcolsep}{3.5pt} 
    \caption{\textbf{Quantitative Study Results.} We evaluate CAdam on representative frameworks corresponding to each objective: GaussianDreamer (SDS),  LucidDreamer (ISM), and FlowDreamer (VFDS).}
    \Description{A quantitative comparison of baseline densification and CAdam under SDS, ISM, and VFDS objectives. The table reports CLIP, ImageReward, HPS v2, NIQE, Gaussian count, and storage size, showing large reductions in Gaussian count and storage for CAdam with broadly comparable quality metrics.}
    \label{tab:quantitative}
    \resizebox{\linewidth}{!}{%
    \begin{tabular}{l|l|cccc|cc}
    \hline
    \multirow{2}{*}{\textbf{Objective}} & \multirow{2}{*}{\textbf{Method}} & \multicolumn{4}{c|}{\textbf{Quality Metrics} $\uparrow$} & \multicolumn{2}{c}{\textbf{Efficiency} $\downarrow$} \\
    \cline{3-8}
     & & CLIP & ImgRwd & HPS v2 & NIQE ($\downarrow$) & \# Gaussians & Storage \\
    \hline
    \hline
    
    \multirow{2}{*}{\textbf{SDS}} & Baseline & 30.14 & -0.125 & 0.194 & 9.908 & 1067K & 46.4MB \\
    \cline{2-8}
     & \textbf{CAdam} & 29.95 & -0.028 & 0.185 & 9.643 & \textbf{143K} & \textbf{6.1MB} \\
    \hline
    
    \multirow{2}{*}{\textbf{ISM}} & Baseline & 31.10 & 0.413 & 0.225 & 6.438 & 1706K & 110.6MB \\
    \cline{2-8}
     & \textbf{CAdam} & 28.48 & 0.518 & 0.225 & 8.167 & \textbf{81K} & \textbf{5.3MB} \\
    \hline
    
    \multirow{2}{*}{\textbf{VFDS}} & Baseline & 29.88 & 0.436 & 0.235 & 7.183 & 3373K & 218.8MB \\
    \cline{2-8}
     & \textbf{CAdam} & 29.21 & 0.486 & 0.237  & 8.891 & \textbf{82K} & \textbf{5.3MB} \\
    \hline
    \end{tabular}%
    }
\end{table}

\paragraph{Human Perception and Preference}
We further assess perceptual quality via a blinded 2AFC study with 30 participants, comparing rendered videos across all frameworks. 
As shown in Tab.~\ref{tab:user_study}, CAdam achieves preference rates comparable to the dense baseline.
The near-balanced overall preference rate of 49.5\% for CAdam versus 50.4\% for the baseline supports our main claim: CAdam achieves substantial compactness gains while maintaining comparable perceived quality.

\begin{table}[t]
    \centering
    \renewcommand{\arraystretch}{1.2}
    \setlength{\tabcolsep}{8pt}
    
    \caption{\textbf{User Study Results.} User preference rates (\%) in a 2AFC setting. Results are averaged across samples from GaussianDreamer, LucidDreamer, and FlowDreamer. Metrics include Text Alignment (\textbf{TA}), Texture Detail (\textbf{TD}), Geometric Plausibility (\textbf{GP}), Cleanliness (\textbf{CL}), and Overall Preference (\textbf{OP}).}
    \Description{A user study table reporting two-alternative forced-choice preference rates for baseline densification and CAdam. The columns compare text alignment, texture detail, geometric plausibility, cleanliness, and overall preference.}
    \label{tab:user_study}
    
    \resizebox{\linewidth}{!}{%
    \begin{tabular}{lccccc}
    \toprule
    \textbf{Method} & \textbf{TA} & \textbf{TD} & \textbf{GP} & \textbf{CL} & \textbf{OP} \\
    \midrule
    Baseline & 50.5\% & 45.8\% & 51.9\% & 55.7\% & 50.4\% \\
    \textbf{CAdam (Ours)} & 49.4\% & 54.1\% & 48.0\% & 44.3\% & 49.5\% \\
    \bottomrule
    \end{tabular}%
    }
\end{table}

\subsection{Analysis of Training Dynamics}
\label{subsec:exp_dynamics}
\begin{figure*}[t]
    \centering
    \includegraphics[width=.96\linewidth]{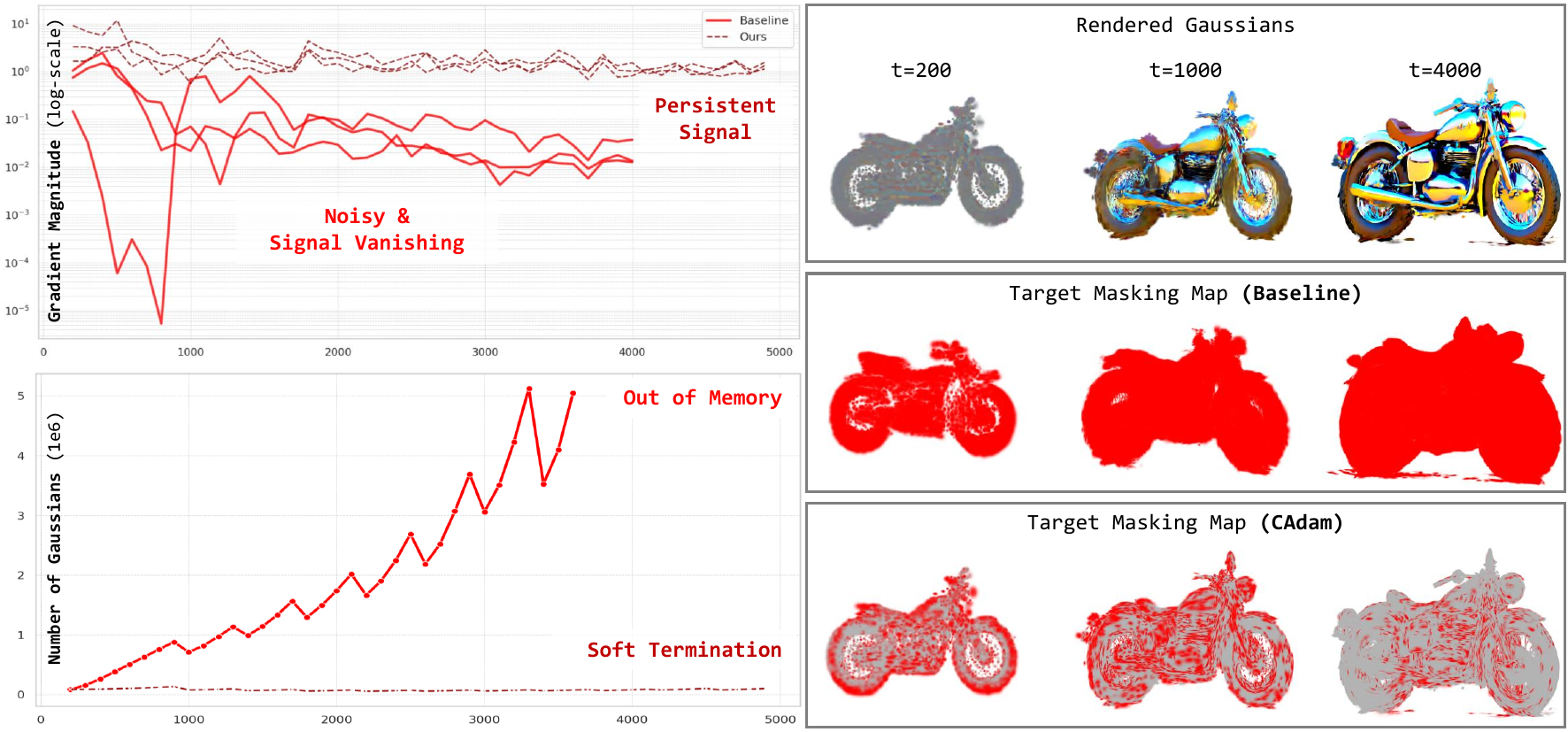}
    \caption{Training dynamics under stochastic generative supervision. Top-left: Gradient magnitude analysis (log scale), tracked for a representative surviving Gaussian, shows that CAdam's momentum-based signal verification maintains persistent structural magnitude to preserve history. Bottom-left: This signal disentanglement directly governs density evolution. Right: Visualization of densification candidates shows that CAdam selectively concentrates refinement on regions of genuine geometric deficiency.}
    \Description{A composite visualization of training dynamics for a motorcycle example. The left side shows gradient magnitude curves and Gaussian count curves, where the baseline exhibits noisy signals and out-of-memory growth while CAdam reaches soft termination. The right side shows rendered Gaussians over time and target masking maps for the baseline and CAdam.}
    \label{fig:6_ex_dynamics}
\end{figure*}

Fig.~\ref{fig:6_ex_dynamics} validates the effectiveness of CAdam's signal verification.
Gradient analysis (top-left) shows that standard accumulation is dominated by noisy fluctuations and decays toward the noise floor, whereas CAdam preserves a more persistent structural signal through momentum-based verification. The primitive-count curve (bottom-left) shows that the baseline exhibits explosive growth and eventually reaches out-of-memory (OOM), while CAdam saturates at a much smaller count. The target masking maps (right) show a transition from coarse shape-level activation to more selective detail-oriented activation, supporting CAdam's soft termination behavior.

\subsection{Ablation Study}
\label{subsec:exp_ablation}

\begin{figure}[t]
    \centering
    \includegraphics[width=.96\linewidth]{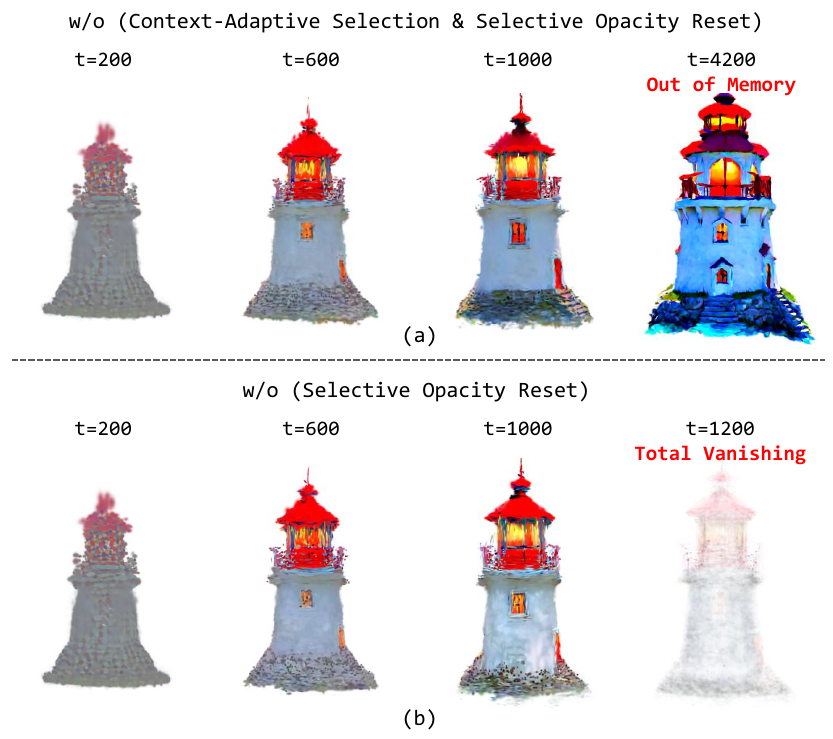}
    \caption{Ablation study using the prompt ``lighthouse, full view, smooth white cylindrical tower, detailed red lantern room, HDR, photorealistic, 8K.'' (a) A momentum-only criterion cannot regulate primitive growth and eventually reaches OOM. (b) Removing Selective Opacity Reset causes progressive over-suppression and eventual vanishing.}
    \Description{Two lighthouse ablation sequences over training time. Without context-adaptive selection and selective opacity reset, the representation keeps growing and reaches out of memory. Without selective opacity reset, the lighthouse progressively fades and eventually vanishes.}
    \label{fig:7_ex_ablation}
\end{figure}

We analyze the contribution of each component in CAdam and its sensitivity to the quantile and SNR thresholds (Figs.~\ref{fig:7_ex_ablation} and \ref{fig:8_ex_threshold}).

\paragraph{Impact of Individual Components}
Fig.~\ref{fig:7_ex_ablation} shows two representative failure modes that reveal the cumulative role of CAdam's components.
A momentum-only criterion suppresses part of the stochastic variation but fails to regulate late-stage primitive growth, eventually leading to OOM.
Adding \textbf{Context-Adaptive Selection} reduces this growth, but without \textbf{Selective Opacity Reset} the representation becomes progressively over-suppressed and eventually vanishes.
These results show that \textbf{Momentum-based Signal Verification} stabilizes noisy updates, \textbf{Context-Adaptive Selection} filters candidates by relative importance and reliability, and \textbf{Selective Opacity Reset} yields the most compact stable configuration by suppressing only low-SNR primitives.

\paragraph{Sensitivity and Controllability}
Fig.~\ref{fig:8_ex_threshold} shows three representative threshold regimes: under-densification, balanced densification, and over-densification.
Higher thresholds act as an aggressive filter and can under-densify weakly supported fine structures, while lower thresholds admit excessive primitive growth and denser representations.
The default setting provides a balanced operating point in our experiments, indicating that the interplay of $\tau_Q$ and $\tau_{\text{SNR}}$ offers a practical control knob over the compactness-quality trade-off.

\section{Conclusion}
\label{sec:conclusion}
Optimization-based generative 3DGS has advanced rapidly, but its densification remains tied to deterministic reconstruction. Under stochastic generative supervision, magnitude accumulation conflates noise with geometric deficiency---the \textit{Densification Dilemma}---causing over-growth or under-refinement. \textbf{CAdam} reframes densification as statistical decision-making: momentum verifies coherent structural signals, while context-adaptive quantile and intrinsic-SNR gates select reliable candidates, yielding self-regulating densification that terminates as structural signals vanish.

\begin{figure}[t]
    \centering
    \includegraphics[width=.96\linewidth]{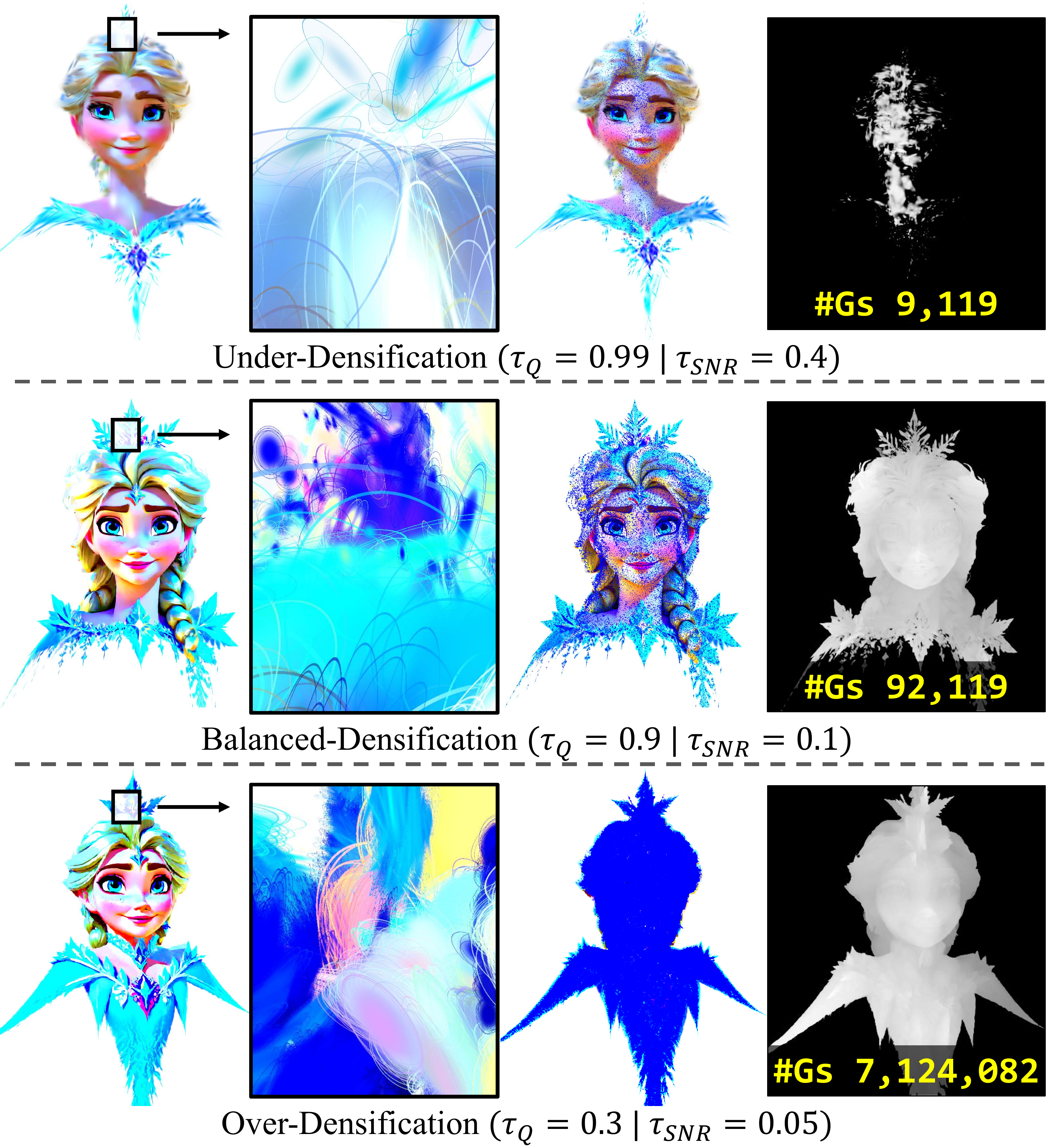}
    \caption{Sensitivity analysis of the quantile and SNR thresholds. The rows show three representative regimes: under-densification, balanced densification, and over-densification. From left to right, we show the rendering, a zoomed-in edge visualization of the boxed region, the point-cloud density map, and the depth map with the final Gaussian count.}
    \Description{Three threshold regimes for a stylized character bust: under-densification, balanced densification, and over-densification. Each row shows a rendering, a zoomed-in edge region, a point-cloud density map, a depth map, and the final Gaussian count.}
    \label{fig:8_ex_threshold}
\end{figure}

Across diverse objectives and backbones, CAdam reduces Gaussian primitives by 85\%--97\% while maintaining overall comparable perceptual quality.
These results suggest that a substantial portion of primitive growth in optimization-based generative 3DGS can be reduced by aligning density evolution with the statistics of generative supervision.
Note that CAdam is not threshold-free: as illustrated in Fig.~\ref{fig:8_ex_threshold}, overly loose thresholds may approach baseline-style over-densification, while overly aggressive thresholds can under-densify weakly supported fine structures such as thin parts or low-contrast details. In our experiments, the default setting provides a stable operating regime across diverse scenarios; remaining failures mainly arise when the underlying generative backbone provides inconsistent semantic or multi-view signals, which is orthogonal to our focus on density control during optimization.

\begin{acks}
This work was supported by the National Research Foundation of Korea (NRF) grants funded by the Korean government (MSIT) (No. RS-2025-00518643 and No. RS-2025-24802983), by the ICT Creative Consilience Program through the Institute of Information \& Communications Technology Planning \& Evaluation (IITP) grant funded by the Korean government (MSIT) (No. IITP-2025-RS-2020-II201819), and by the Institute of Information \& communications Technology Planning \& Evaluation (IITP) grant funded by the Korea government (MSIT) (No. RS-2020-II200861).
\end{acks}

\clearpage

\bibliographystyle{ACM-Reference-Format}
\bibliography{main-sigconf}

@String{Computing = "Computing" }

@String{Computer = "{IEEE} Computer" }

@String{Springer = "Springer-Verlag" }

@inproceedings{rombach2022high,
  title={High-resolution image synthesis with latent diffusion models},
  author={Rombach, Robin and Blattmann, Andreas and Lorenz, Dominik and Esser, Patrick and Ommer, Bj{\"o}rn},
  booktitle={Proceedings of the IEEE/CVF conference on computer vision and pattern recognition},
  pages={10684--10695},
  year={2022}
}

@article{lipman2022flow,
  title={Flow matching for generative modeling},
  author={Lipman, Yaron and Chen, Ricky TQ and Ben-Hamu, Heli and Nickel, Maximilian and Le, Matt},
  journal={arXiv preprint arXiv:2210.02747},
  year={2022}
}

@article{liu2022flow,
  title={Flow straight and fast: Learning to generate and transfer data with rectified flow},
  author={Liu, Xingchao and Gong, Chengyue and Liu, Qiang},
  journal={arXiv preprint arXiv:2209.03003},
  year={2022}
}

@article{li2024flowdreamer,
  title={Flowdreamer: Exploring high fidelity text-to-3d generation via rectified flow},
  author={Li, Hangyu and Chu, Xiangxiang and Shi, Dingyuan and Lin, Wang},
  journal={arXiv preprint arXiv:2408.05008},
  year={2024}
}

@article{kerbl20233d,
  title={3D Gaussian splatting for real-time radiance field rendering.},
  author={Kerbl, Bernhard and Kopanas, Georgios and Leimk{\"u}hler, Thomas and Drettakis, George},
  journal={ACM Trans. Graph.},
  volume={42},
  number={4},
  pages={139--1},
  year={2023}
}

@inproceedings{
poole2023dreamfusion,
title={DreamFusion: Text-to-3D using 2D Diffusion},
author={Ben Poole and Ajay Jain and Jonathan T. Barron and Ben Mildenhall},
booktitle={The Eleventh International Conference on Learning Representations },
year={2023},
url={https://openreview.net/forum?id=FjNys5c7VyY}
}

@inproceedings{alldieck2024score,
  title={Score distillation sampling with learned manifold corrective},
  author={Alldieck, Thiemo and Kolotouros, Nikos and Sminchisescu, Cristian},
  booktitle={European Conference on Computer Vision},
  pages={1--18},
  year={2024},
  organization={Springer}
}

@article{chen2024vividdreamer,
  title={VividDreamer: Towards High-Fidelity and Efficient Text-to-3D Generation},
  author={Chen, Zixuan and Su, Ruijie and Zhu, Jiahao and Yang, Lingxiao and Lai, Jian-Huang and Xie, Xiaohua},
  journal={arXiv preprint arXiv:2406.14964},
  year={2024}
}

@inproceedings{li2024connecting,
  title={Connecting consistency distillation to score distillation for text-to-3d generation},
  author={Li, Zongrui and Hu, Minghui and Zheng, Qian and Jiang, Xudong},
  booktitle={European Conference on Computer Vision},
  pages={274--291},
  year={2024},
  organization={Springer}
}

@inproceedings{zhang2024pixel,
  title={Pixel-gs: Density control with pixel-aware gradient for 3d gaussian splatting},
  author={Zhang, Zheng and Hu, Wenbo and Lao, Yixing and He, Tong and Zhao, Hengshuang},
  booktitle={European Conference on Computer Vision},
  pages={326--342},
  year={2024},
  organization={Springer}
}

@inproceedings{ye2024absgs,
  title={Absgs: Recovering fine details in 3d gaussian splatting},
  author={Ye, Zongxin and Li, Wenyu and Liu, Sidun and Qiao, Peng and Dou, Yong},
  booktitle={Proceedings of the 32nd ACM International Conference on Multimedia},
  pages={1053--1061},
  year={2024}
}

@article{fan2024trim,
  title={Trim 3d gaussian splatting for accurate geometry representation},
  author={Fan, Lue and Yang, Yuxue and Li, Minxing and Li, Hongsheng and Zhang, Zhaoxiang},
  journal={arXiv preprint arXiv:2406.07499},
  year={2024}
}

@article{kotovenko2025edgs,
  title={EDGS: Eliminating Densification for Efficient Convergence of 3DGS},
  author={Kotovenko, Dmytro and Grebenkova, Olga and Ommer, Bj{\"o}rn},
  journal={arXiv preprint arXiv:2504.13204},
  year={2025}
}

@inproceedings{lyu2025resgs,
  title={Resgs: Residual densification of 3d gaussian for efficient detail recovery},
  author={Lyu, Yanzhe and Cheng, Kai and Kang, Xin and Chen, Xuejin},
  booktitle={Proceedings of the IEEE/CVF International Conference on Computer Vision},
  pages={28093--28102},
  year={2025}
}

@article{yuan2025ema,
  title={EMA-GS: Improving sparse point cloud rendering with EMA gradient and anchor upsampling},
  author={Yuan, Ding and Zhang, Sizhe and Zhang, Hong and Deng, Yangyan and Yang, Yifan},
  journal={Image and Vision Computing},
  volume={154},
  pages={105433},
  year={2025},
  publisher={Elsevier}
}

@article{kheradmand20243d,
  title={3d gaussian splatting as markov chain monte carlo},
  author={Kheradmand, Shakiba and Rebain, Daniel and Sharma, Gopal and Sun, Weiwei and Tseng, Yang-Che and Isack, Hossam and Kar, Abhishek and Tagliasacchi, Andrea and Yi, Kwang Moo},
  journal={Advances in Neural Information Processing Systems},
  volume={37},
  pages={80965--80986},
  year={2024}
}

@inproceedings{kim2024color,
  title={Color-cued efficient densification method for 3d gaussian splatting},
  author={Kim, Sieun and Lee, Kyungjin and Lee, Youngki},
  booktitle={Proceedings of the IEEE/CVF Conference on Computer Vision and Pattern Recognition},
  pages={775--783},
  year={2024}
}

@article{li2025psrgs,
  title={PSRGS: Progressive Spectral Residual of 3D Gaussian for High-Frequency Recovery},
  author={Li, BoCheng and Zhang, WenJuan and Zhang, Bing and Yao, YiLing and Wang, YaNing},
  journal={arXiv preprint arXiv:2503.00848},
  year={2025}
}

@article{ye20243d,
  title={3D Gaussian rendering can be sparser: Efficient rendering via learned fragment pruning},
  author={Ye, Zhifan and Wan, Chenxi and Li, Chaojian and Hong, Jihoon and Li, Sixu and Li, Leshu and Zhang, Yongan and Lin, Yingyan Celine},
  journal={Advances in Neural Information Processing Systems},
  volume={37},
  pages={5850--5869},
  year={2024}
}

@inproceedings{kingma2015adam,
  title={Adam: A Method for Stochastic Optimization},
  author={Kingma, Diederik P. and Ba, Jimmy},
  booktitle={3rd International Conference on Learning Representations (ICLR)},
  year={2015},
  address={San Diego, CA, USA}
}

@article{zhu2023hifa,
  title={Hifa: High-fidelity text-to-3d generation with advanced diffusion guidance},
  author={Zhu, Junzhe and Zhuang, Peiye and Koyejo, Sanmi},
  journal={arXiv preprint arXiv:2305.18766},
  year={2023}
}

@inproceedings{wang2023score,
  title={Score jacobian chaining: Lifting pretrained 2d diffusion models for 3d generation},
  author={Wang, Haochen and Du, Xiaodan and Li, Jiahao and Yeh, Raymond A and Shakhnarovich, Greg},
  booktitle={Proceedings of the IEEE/CVF conference on computer vision and pattern recognition},
  pages={12619--12629},
  year={2023}
}

@article{yu2023text,
  title={Text-to-3d with classifier score distillation},
  author={Yu, Xin and Guo, Yuan-Chen and Li, Yangguang and Liang, Ding and Zhang, Song-Hai and Qi, Xiaojuan},
  journal={arXiv preprint arXiv:2310.19415},
  year={2023}
}

@article{wang2023prolificdreamer,
  title={Prolificdreamer: High-fidelity and diverse text-to-3d generation with variational score distillation},
  author={Wang, Zhengyi and Lu, Cheng and Wang, Yikai and Bao, Fan and Li, Chongxuan and Su, Hang and Zhu, Jun},
  journal={Advances in neural information processing systems},
  volume={36},
  pages={8406--8441},
  year={2023}
}

@inproceedings{wu2024consistent3d,
  title={Consistent3d: Towards consistent high-fidelity text-to-3d generation with deterministic sampling prior},
  author={Wu, Zike and Zhou, Pan and Yi, Xuanyu and Yuan, Xiaoding and Zhang, Hanwang},
  booktitle={Proceedings of the IEEE/CVF Conference on Computer Vision and Pattern Recognition},
  pages={9892--9902},
  year={2024}
}

@article{yan2025consistent,
  title={Consistent flow distillation for text-to-3d generation},
  author={Yan, Runjie and Chen, Yinbo and Wang, Xiaolong},
  journal={arXiv preprint arXiv:2501.05445},
  year={2025}
}

@inproceedings{yi2024gaussiandreamer,
  title={Gaussiandreamer: Fast generation from text to 3d gaussians by bridging 2d and 3d diffusion models},
  author={Yi, Taoran and Fang, Jiemin and Wang, Junjie and Wu, Guanjun and Xie, Lingxi and Zhang, Xiaopeng and Liu, Wenyu and Tian, Qi and Wang, Xinggang},
  booktitle={Proceedings of the IEEE/CVF Conference on Computer Vision and Pattern Recognition},
  pages={6796--6807},
  year={2024}
}

@article{tang2023dreamgaussian,
  title={Dreamgaussian: Generative gaussian splatting for efficient 3d content creation},
  author={Tang, Jiaxiang and Ren, Jiawei and Zhou, Hang and Liu, Ziwei and Zeng, Gang},
  journal={arXiv preprint arXiv:2309.16653},
  year={2023}
}

@inproceedings{chen2024text,
  title={Text-to-3d using gaussian splatting},
  author={Chen, Zilong and Wang, Feng and Wang, Yikai and Liu, Huaping},
  booktitle={Proceedings of the IEEE/CVF conference on computer vision and pattern recognition},
  pages={21401--21412},
  year={2024}
}

@inproceedings{liang2024luciddreamer,
  title={Luciddreamer: Towards high-fidelity text-to-3d generation via interval score matching},
  author={Liang, Yixun and Yang, Xin and Lin, Jiantao and Li, Haodong and Xu, Xiaogang and Chen, Yingcong},
  booktitle={Proceedings of the IEEE/CVF conference on computer vision and pattern recognition},
  pages={6517--6526},
  year={2024}
}

@inproceedings{zhong2024dreamlcm,
  title={Dreamlcm: Towards high quality text-to-3d generation via latent consistency model},
  author={Zhong, Yiming and Zhang, Xiaolin and Zhao, Yao and Wei, Yunchao},
  booktitle={Proceedings of the 32nd ACM International Conference on Multimedia},
  pages={1731--1740},
  year={2024}
}

@article{miao2024dreamer,
  title={Dreamer XL: Towards High-Resolution Text-to-3D Generation via Trajectory Score Matching},
  author={Miao, Xingyu and Duan, Haoran and Ojha, Varun and Song, Jun and Shah, Tejal and Long, Yang and Ranjan, Rajiv},
  journal={arXiv preprint arXiv:2405.11252},
  year={2024}
}

@article{lukoianov2024score,
  title={Score distillation via reparametrized ddim},
  author={Lukoianov, Artem and S{\'a}ez de Oc{\'a}riz Borde, Haitz and Greenewald, Kristjan and Guizilini, Vitor and Bagautdinov, Timur and Sitzmann, Vincent and Solomon, Justin M},
  journal={Advances in Neural Information Processing Systems},
  volume={37},
  pages={26011--26044},
  year={2024}
}

@article{li2025walking,
  title={Walking the Schr{\"o}dinger Bridge: A Direct Trajectory for Text-to-3D Generation},
  author={Li, Ziying and Lu, Xuequan and Zhao, Xinkui and Cheng, Guanjie and Deng, Shuiguang and Yin, Jianwei},
  journal={arXiv preprint arXiv:2511.05609},
  year={2025}
}

@article{zhou2025consdreamer,
  title={ConsDreamer: Advancing Multi-View Consistency for Zero-Shot Text-to-3D Generation},
  author={Zhou, Yuan and Jin, Shilong and Hua, Litao and Lv, Wanjun and Duan, Haoran and Han, Jungong},
  journal={arXiv preprint arXiv:2504.02316},
  year={2025}
}

@article{jin2025debiasing,
  title={Debiasing Diffusion Priors via 3D Attention for Consistent Gaussian Splatting},
  author={Jin, Shilong and Duan, Haoran and Hua, Litao and Huang, Wentao and Zhou, Yuan},
  journal={arXiv preprint arXiv:2512.07345},
  year={2025}
}

@inproceedings{zhang2025improving,
  title={Improving Viewpoint Consistency in 3D Generation via Structure Feature and CLIP Guidance},
  author={Zhang, Qing and Tong, Jinguang and Zhang, Jing and Hong, Jie and Li, Xuesong},
  booktitle={Proceedings of the IEEE/CVF International Conference on Computer Vision},
  pages={6440--6449},
  year={2025}
}

@inproceedings{zhu2025segmentdreamer,
  title={SegmentDreamer: Towards High-fidelity Text-to-3D Synthesis with Segmented Consistency Trajectory Distillation},
  author={Zhu, Jiahao and Chen, Zixuan and Wang, Guangcong and Xie, Xiaohua and Zhou, Yi},
  booktitle={Proceedings of the IEEE/CVF International Conference on Computer Vision},
  pages={15864--15874},
  year={2025}
}

@article{zhang2025structural,
  title={Structural Energy-Guided Sampling for View-Consistent Text-to-3D},
  author={Zhang, Qing and Tong, Jinguang and Hong, Jie and Zhang, Jing and Li, Xuesong},
  journal={arXiv preprint arXiv:2508.16917},
  year={2025}
}

@article{cai2024dreammapping,
  title={DreamMapping: High-Fidelity Text-to-3D Generation via Variational Distribution Mapping},
  author={Cai, Zeyu and Wang, Duotun and Liang, Yixun and Shao, Zhijing and Chen, Ying-Cong and Zhan, Xiaohang and Wang, Zeyu},
  journal={arXiv preprint arXiv:2409.05099},
  year={2024}
}

@inproceedings{he2024gvgen,
  title={Gvgen: Text-to-3d generation with volumetric representation},
  author={He, Xianglong and Chen, Junyi and Peng, Sida and Huang, Di and Li, Yangguang and Huang, Xiaoshui and Yuan, Chun and Ouyang, Wanli and He, Tong},
  booktitle={European Conference on Computer Vision},
  pages={463--479},
  year={2024},
  organization={Springer}
}

@article{di2025hyper,
  title={Hyper-3dg: Text-to-3d gaussian generation via hypergraph},
  author={Di, Donglin and Yang, Jiahui and Luo, Chaofan and Xue, Zhou and Chen, Wei and Yang, Xun and Gao, Yue},
  journal={International Journal of Computer Vision},
  volume={133},
  number={5},
  pages={2886--2909},
  year={2025},
  publisher={Springer}
}

@article{pham2024mvgaussian,
  title={Mvgaussian: High-fidelity text-to-3d content generation with multi-view guidance and surface densification},
  author={Pham, Phu and Mathur, Aradhya N and Sharma, Ojaswa and Bera, Aniket},
  journal={arXiv preprint arXiv:2409.06620},
  year={2024}
}

@inproceedings{huang2024placiddreamer,
  title={Placiddreamer: Advancing harmony in text-to-3d generation},
  author={Huang, Shuo and Sun, Shikun and Wang, Zixuan and Qin, Xiaoyu and Xiong, Yanmin and Zhang, Yuan and Wan, Pengfei and Zhang, Di and Jia, Jia},
  booktitle={Proceedings of the 32nd ACM International Conference on Multimedia},
  pages={6880--6889},
  year={2024}
}

@article{tran2024modedreamer,
  title={ModeDreamer: Mode Guiding Score Distillation for Text-to-3D Generation using Reference Image Prompts},
  author={Tran, Uy Dieu and Luu, Minh and Nguyen, Phong Ha and Nguyen, Khoi and Hua, Binh-Son},
  journal={arXiv preprint arXiv:2411.18135},
  year={2024}
}

@article{miao2025rethinking,
  title={Rethinking Score Distilling Sampling for 3D Editing and Generation},
  author={Miao, Xingyu and Duan, Haoran and Long, Yang and Han, Jungong},
  journal={arXiv preprint arXiv:2505.01888},
  year={2025}
}

@inproceedings{zhuo2024vividdreamer,
  title={Vividdreamer: invariant score distillation for hyper-realistic text-to-3d generation},
  author={Zhuo, Wenjie and Ma, Fan and Fan, Hehe and Yang, Yi},
  booktitle={European Conference on Computer Vision},
  pages={122--139},
  year={2024},
  organization={Springer}
}

@inproceedings{li2025controllable,
  title={Controllable text-to-3D generation via surface-aligned Gaussian splatting},
  author={Li, Zhiqi and Chen, Yiming and Zhao, Lingzhe and Liu, Peidong},
  booktitle={2025 International Conference on 3D Vision (3DV)},
  pages={1113--1123},
  year={2025},
  organization={IEEE}
}

@inproceedings{tang2025cycle3d,
  title={Cycle3d: High-quality and consistent image-to-3d generation via generation-reconstruction cycle},
  author={Tang, Zhenyu and Zhang, Junwu and Cheng, Xinhua and Yu, Wangbo and Feng, Chaoran and Pang, Yatian and Lin, Bin and Yuan, Li},
  booktitle={Proceedings of the AAAI Conference on Artificial Intelligence},
  volume={39},
  number={7},
  pages={7320--7328},
  year={2025}
}

@inproceedings{xu2024grm,
  title={Grm: Large gaussian reconstruction model for efficient 3d reconstruction and generation},
  author={Xu, Yinghao and Shi, Zifan and Yifan, Wang and Chen, Hansheng and Yang, Ceyuan and Peng, Sida and Shen, Yujun and Wetzstein, Gordon},
  booktitle={European Conference on Computer Vision},
  pages={1--20},
  year={2024},
  organization={Springer}
}

@inproceedings{zhang2024gs,
  title={Gs-lrm: Large reconstruction model for 3d gaussian splatting},
  author={Zhang, Kai and Bi, Sai and Tan, Hao and Xiangli, Yuanbo and Zhao, Nanxuan and Sunkavalli, Kalyan and Xu, Zexiang},
  booktitle={European Conference on Computer Vision},
  pages={1--19},
  year={2024},
  organization={Springer}
}

@inproceedings{tang2024lgm,
  title={Lgm: Large multi-view gaussian model for high-resolution 3d content creation},
  author={Tang, Jiaxiang and Chen, Zhaoxi and Chen, Xiaokang and Wang, Tengfei and Zeng, Gang and Liu, Ziwei},
  booktitle={European Conference on Computer Vision},
  pages={1--18},
  year={2024},
  organization={Springer}
}

@article{liu2024novelgs,
  title={Novelgs: Consistent novel-view denoising via large gaussian reconstruction model},
  author={Liu, Jinpeng and Xu, Jiale and Cheng, Weihao and Gao, Yiming and Wang, Xintao and Shan, Ying and Tang, Yansong},
  journal={arXiv preprint arXiv:2411.16779},
  year={2024}
}

@inproceedings{hu2025turbo3d,
  title={Turbo3d: Ultra-fast text-to-3d generation},
  author={Hu, Hanzhe and Yin, Tianwei and Luan, Fujun and Hu, Yiwei and Tan, Hao and Xu, Zexiang and Bi, Sai and Tulsiani, Shubham and Zhang, Kai},
  booktitle={Proceedings of the Computer Vision and Pattern Recognition Conference},
  pages={23668--23678},
  year={2025}
}

@article{li2025back,
  title={Back to basics: Let denoising generative models denoise},
  author={Li, Tianhong and He, Kaiming},
  journal={arXiv preprint arXiv:2511.13720},
  year={2025}
}

@misc{he2023t3bench,
    title={T$^3$Bench: Benchmarking Current Progress in Text-to-3D Generation}, 
    author={Yuze He and Yushi Bai and Matthieu Lin and Wang Zhao and Yubin Hu and Jenny Sheng and Ran Yi and Juanzi Li and Yong-Jin Liu},
    year={2023},
    eprint={2310.02977},
    archivePrefix={arXiv},
    primaryClass={cs.CV}
}

@inproceedings{hessel2021clipscore,
  title={Clipscore: A reference-free evaluation metric for image captioning},
  author={Hessel, Jack and Holtzman, Ari and Forbes, Maxwell and Le Bras, Ronan and Choi, Yejin},
  booktitle={Proceedings of the 2021 conference on empirical methods in natural language processing},
  pages={7514--7528},
  year={2021}
}

@article{xu2023imagereward,
  title={Imagereward: Learning and evaluating human preferences for text-to-image generation},
  author={Xu, Jiazheng and Liu, Xiao and Wu, Yuchen and Tong, Yuxuan and Li, Qinkai and Ding, Ming and Tang, Jie and Dong, Yuxiao},
  journal={Advances in Neural Information Processing Systems},
  volume={36},
  pages={15903--15935},
  year={2023}
}

@article{wu2023human,
  title={Human Preference Score v2: A Solid Benchmark for Evaluating Human Preferences of Text-to-Image Synthesis},
  author={Wu, Xiaoshi and Hao, Yiming and Sun, Keqiang and Chen, Yixiong and Zhu, Feng and Zhao, Rui and Li, Hongsheng},
  journal={arXiv preprint arXiv:2306.09341},
  year={2023}
}

@article{mittal2012making,
  title={Making a ``completely blind'' image quality analyzer},
  author={Mittal, Anish and Soundararajan, Rajiv and Bovik, Alan C},
  journal={IEEE Signal processing letters},
  volume={20},
  number={3},
  pages={209--212},
  year={2012},
  publisher={IEEE}
}

@article{xiang2024structured,
    title   = {Structured 3D Latents for Scalable and Versatile 3D Generation},
    author  = {Xiang, Jianfeng and Lv, Zelong and Xu, Sicheng and Deng, Yu and Wang, Ruicheng and Zhang, Bowen and Chen, Dong and Tong, Xin and Yang, Jiaolong},
    journal = {arXiv preprint arXiv:2412.01506},
    year    = {2024}
}

@article{
    xiang2025trellis2,
    title={Native and Compact Structured Latents for 3D Generation},
    author={Xiang, Jianfeng and Chen, Xiaoxue and Xu, Sicheng and Wang, Ruicheng and Lv, Zelong and Deng, Yu and Zhu, Hongyuan and Dong, Yue and Zhao, Hao and Yuan, Nicholas Jing and Yang, Jiaolong},
    journal={Tech report},
    year={2025}
}

@article{wu2025unilat3d,
  title={Unilat3d: Geometry-appearance unified latents for single-stage 3d generation},
  author={Wu, Guanjun and Fang, Jiemin and Yang, Chen and Li, Sikuang and Yi, Taoran and Lu, Jia and Zhou, Zanwei and Cen, Jiazhong and Xie, Lingxi and Zhang, Xiaopeng and others},
  journal={arXiv preprint arXiv:2509.25079},
  year={2025}
}

@article{lai2025hunyuan3d,
  title={Hunyuan3d 2.5: Towards high-fidelity 3d assets generation with ultimate details},
  author={Lai, Zeqiang and Zhao, Yunfei and Liu, Haolin and Zhao, Zibo and Lin, Qingxiang and Shi, Huiwen and Yang, Xianghui and Yang, Mingxin and Yang, Shuhui and Feng, Yifei and others},
  journal={arXiv preprint arXiv:2506.16504},
  year={2025}
}

@inproceedings{hanson2025pup,
  title={Pup 3d-gs: Principled uncertainty pruning for 3d gaussian splatting},
  author={Hanson, Alex and Tu, Allen and Singla, Vasu and Jayawardhana, Mayuka and Zwicker, Matthias and Goldstein, Tom},
  booktitle={Proceedings of the Computer Vision and Pattern Recognition Conference},
  pages={5949--5958},
  year={2025}
}

@article{zhang2024lp,
  title={Lp-3dgs: Learning to prune 3d gaussian splatting},
  author={Zhang, Zhaoliang and Song, Tianchen and Lee, Yongjae and Yang, Li and Peng, Cheng and Chellappa, Rama and Fan, Deliang},
  journal={Advances in Neural Information Processing Systems},
  volume={37},
  pages={122434--122457},
  year={2024}
}

\end{document}